# Optimal Limited Contingency Planning

Nicolas Meuleau* and David E. Smith
NASA Ames Research Center
Mail Stop 269-3
Moffet Field, CA 94035-1000
{nmeuleau, de2smith}@email.arc.nasa.gov

## Abstract

For a given problem, the optimal Markov policy over a finite horizon is a conditional plan containing a potentially large number of branches. However, there are applications where it is desirable to strictly limit the number of decision points and branches in a plan. This raises the question of how one goes about finding optimal plans containing only a limited number of branches. In this paper, we present an any-time algorithm for optimal $k$-contingency planning. It is the first optimal algorithm for limited contingency planning that is not an explicit enumeration of possible contingent plans. By modelling the problem as a partially observable Markov decision process, it implements the Bellman optimality principle and prunes the solution space. We present experimental results of applying this algorithm to some simple test cases.

## 1 INTRODUCTION

*Markov decision processes* (MDPs) provide a powerful theoretical framework for planning under uncertainty with probabilities, costs and rewards [15]. In this framework, the optimal solution to a problem is an optimal *policy*, that is, a rule specifying the action to perform for each situation we could possibly be in. For a finite planning horizon, this policy represents a *conditional* or *contingent* plan with a branch for each possible situation that might be encountered during execution. Therefore, the optimal contingent plan may be large and complex, since it may contain a large number of branches.

There are applications where this size and complexity is a significant drawback. Consider, for example, the problem of constructing daily plans for a Mars rover. There is a great deal of uncertainty in this domain, concerning such things

*QSS Group Inc.

as time, energy usage, data storage available, and position (see [5] for a more detailed description). However, there are some compelling reasons for keeping the plans simple:

- There is a need for cognitive simplicity – plans must be simple enough that they can be displayed easily, and understood and modified by both Earth scientists and mission operations personnel.

- Plans must undergo very detailed analysis and simulation using complex models of illumination, energy consumption, thermal characteristics, kinematics, and terrain. There is limited time to do this analysis, so plans must be kept structurally simple in order to expedite this process.

- There is limited communication bandwidth and limited storage on board the rover, so there is an advantage to keeping plans small.

As a result, we are interested in *limited* contingency planning. More precisely, we would like to be able to compute the optimal $k$-contingency plan for a problem – that is, the optimal plan containing $k$ or fewer contingency branches.

In general, the problem of contingency planning is known to be quite hard [11], and $k$-contingency planning is no exception. If $k = \infty$, $k$-contingency planning reduces to finding the optimal policy. If $k = 0$, $k$-contingency planning reduces to stochastic *conformant* planning, where we must find the best unconditional sequence of actions [9]. One can argue that limited contingency planning is harder than both conformant planning and searching for the optimal policy. First, the search space of conformant planing (that is, the set of all sequences of actions) is exponentially smaller than the search space of $k$-contingency planning (the set of all $k$-contingency plans). Second, although the set of all policies is usually larger than the set of all $k$-contingency plans, dynamic programming (DP) techniques are able to significantly prune the search for an optimal policy by using Bellman's optimality principle. However, to our knowledge, there is no previous algorithm that is able to implement Bellman's optimality principle for limited contingency planning. The problem is that the classical Markov state is insufficient: knowing the best limited contingency plan from time $t+1$ to the horizon for each state we could be in at time $t+1$ does not help to find the best plan from time $t$ to the horizon. In fact, the action performed at time $t$ may bring us no certainty about the state at time $t+1$, and the best plan for an uncertain initial state may be different from the best plan in each state. However, the belief-state borrowed from partially observable Markov decision process (POMDP) theory [6, 10], that is, a probability distribution on the original state, is a sufficient statistic to allow a DP approach to the problem of limited contingency planning. This is the basic principle of the algorithm presented in this paper.

Conformant planning is well known to be equivalent to the problem of planning in an unobservable environment: limiting oneself to unconditional plans is equivalent to discarding the observation of the current state that is available at each time step. The first algorithm to exploit this fact performed heuristic search through the belief space [1, 4]. Instead of using Bellman's optimality principle, these techniques (when they tackle the *optimal* planning problem) rely on admissible heuristics to prune the search space [4]. Recently, Hyafil and Bacchus used the best solution techniques for POMDPs to solve stochastic conformant planning problems [9]. In this approach, conformant planning is modelled as a fully *non-observable Markov decision process* (NOMDP), which is a particular case of a POMDP. As Hyafil and Bacchus point out, the drawback of this approach is that it requires computing optimal solutions for states that may be unreachable, but its strength is that it prunes the search space by using Bellman's optimality principle. For several test bed problems, Hyafil and Bacchus show that this approach outperforms a CSP algorithm that is able to do some reachability analysis but cannot prune the search space. Moreover, the superiority of the POMDP approach becomes apparent as the size of the problems grows.

In this paper, we present *optimal k-contingency planning* (OKP), an incremental algorithm for optimal limited contingency planning. As in [9], we use a POMDP framework to model the problem, which allows using Bellman's optimality principle to speed up the search. The difference is that we must encode the number of branches allowed in the state description of the POMDP. In effect, this amounts to keeping multiple copies of the POMDP corresponding to different numbers of branches allowed. When we choose to make an observation in one POMDP, we drop into a POMDP with fewer branches allowed. When all the branches are used up, we end up in the POMDP for the conformant planning problem as used by Hyafil and Bacchus.

We start by specifying the notion of contingent plan used throughout the paper. In Section 2, we first show how Hyafil and Bacchus encoded conformant planning as a POMDP. We then move on to 1-contingency planning, followed by *balanced k-contingency planning*. In Section 3 we further generalize this to arbitrary $k$-contingency planning. In Section 4 we present experimental results comparing OKP against a brute force search technique for finding $k$-contingency plans. Finally, we review related work and conclude.

### 1.1 CONTINGENT PLANS

This paper addresses a series of variants of the limited contingency planning problem. In general, we are looking for optimal tree-shaped plans, the simplest form being conformant plans, which are simple sequences of actions without branches. This choice may seem a little odd since there are more compact types of plans or policies, such as finite state controllers. However, there are reasons to prefer tree-shaped plans in some application domains. For instance, in the Mars rover domain, resources are monotonically decreasing along each possible trajectory, so that a state is never visited twice. Moreover, the action the rover executes must depend on the resource available. Therefore, NASA requires that plans have finite horizon and do not contain loops.

*Optimal k-contingency planning* is the problem of finding an optimal tree-shaped plan with (at most) $k$ branch points. We consider three variants of this problem:

**general $k$-contingency planning:** in the most general case, we are looking for the best plan with at most $k$ branch points overall;

**linear $k$-contingency planning:** we try to find the best plan with at most $k$ branch points, all of them on one trajectory through the plan. That is, the plan structure is a main line of actions with simple branches attached to it, and no branches on the branches.

**balanced $k$-contingency planning:** we are looking for the best plan with at most $k$ branch points in each possible trajectory through the plan. That is, the largest possible plan structure is a balanced tree with $k$ branch points in each path from the root (initial time) to a leaf (planning horizon). So, there are actually more than $k$ branch points over the whole plan.

Although the balanced plan structure is a bit contrived, it is useful for presenting our algorithm since OKP takes its simplest form in this case.

## 2   OPTIMAL BALANCED $k$-CONTINGENCY PLANNING

Our formalism uses several POMDPs defined over different state, action and observation spaces, so it is important to understand the role of each POMDP. The first POMDP we introduce, $M$, represents the planning problem in the classical sense. In this paper, our goal is to find optimal contingent plans for the process $M$. $M$ can be a fully observable MDP, which we see as a particular case of a POMDP. In our framework, it means that we can observe exactly the current state each time we decide to branch. In the general case (when $M$ is not an MDP), we have only noisy observations for branching decisions. Later, we introduce several other POMDPs, $\{M^k : k \geq 0\}$, obtained by transforming the original process $M$ in such a way that an optimal solution to $M^k$ is an optimal $k$-contingency plan for $M$. So, each $M^k$ represents *the problem of $k$-contingency planning in $M$*.

The planning problem for which we want to find optimal contingent plans is modelled as the POMDP $M = (S, A, \Omega, T, R, O)$, where $S$, $A$ and $\Omega$ are the (finite) set of states, actions and observations (respectively); $T$ is the transition probability: $T(s, a, s')$ is the probability of moving to state $s'$ if we execute action $a$ in state $s$; $R$ is the reward function: $R(s, a)$ is the (expected) reward for executing action $a$ in state $s$; and $O$ is the observation probability: $O(a, s', o)$ is the probability of observing $o \in \Omega$ when an execution of action $a$ leads to state $s'$. In this section, we assume that the observation probabilities of $M$ do not depend on the last action executed, and we denote by $O(s', o)$ the (well defined) probability of observing $o \in \Omega$ when arriving in $s' \in S$. We relax this assumption in Section 3.3. If $M$ is a fully-observable MDP, then $\Omega = S$ and $O(s', s') = 1$ for all $s' \in S$.

The problem we tackle is this section is the following: given $M$, $H$, and a probability distribution over the initial state $x_0(s)$ (the initial belief), find the best contingent plan where there are (at most) $k$ branch points in each possible trajectory through the plan. The optimality criterion used is the classical expected cumulative reward (discounted or not) up to the planning horizon $H$:

$$E\left[\sum_{t=1}^{H} \gamma^t r(t) \mid x_0\right] ,$$

$r(t)$ is the reward received at time $t$ and $\gamma \in [0, 1]$ is the discount factor.

First, we assume that we must create one branch for each observation that can be made at each branch point (so, the branch points are $|O|$-ary in a POMDP, and $|S|$-ary in an MDP). We show how to relax this constraint in Section 3.2.

### 2.1 CONFORMANT PLANNING

When $k = 0$, the problem is that of conformant planning: we must find the best unconditional sequence of $H$ actions. As Hyafil and Bacchus [9], we model the stochastic conformant planning problem as a completely non observable MDP (NOMDP) $M^0 = (S^0, A^0, \Omega^0, T^0, R^0, O^0)$ where $S^0 = S$; $A^0 = A$; $\Omega^0$ contains only one element, $o^0$, that basically says *"I can't see anything informative"*; $T^0(s, a, s') = T(s, a, s')$, $R^0(s, a) = R(s, a)$, and $O^0(a, s', o^0) = 1$ for all $(s, a, s') \in S \times A \times S$.

As for any POMDP [10], the optimal solution of $M^0$ over the finite horizon $H$ can be determined in finite time using *value iteration* (VI), which is a form of dynamic programming (DP). Starting from the planning horizon $H$, we proceed backward through time to construct a value function $V_t^0$ for each $t \in \{0; 1; \ldots H\}$. The value $V_t^0(x)$ represents the expected reward we get by executing an optimal conformant plan for the starting belief $x$ over the planning horizon $t$. In the particular case of the NOMDP $M^0$, the equations of VI are the following (the superscript 0 of the $V$ and $Q$ functions is a reference to $k$, the number of branch points in the plan):

$$V_H^0(x) = 0 , \qquad (1)$$

and, for all $t \in \{0, 1, \ldots H - 1\}$:

$$V_t^0(x) = \max_{a \in A} \left[Q_t^0(x, a)\right] , \qquad (2)$$

$$Q_t^0(x, a) = \left(\sum_{s \in S} x(s) R(s, a)\right) + \gamma V_{t+1}^0(\mathcal{B}_{o^0}^a(x)) . \qquad (3)$$

$\mathcal{B}_{o^0}^a(x)$ represents the belief posterior to action $a$ and observation $o^0$, given the prior belief $x$. It is given by Bayes' rule:

$$\mathcal{B}_{o^0}^a(x)(s') = \frac{\sum_{s \in S} x(s) T(s, a, s')}{Z} . \qquad (4)$$

Since we do not make any observation at all, whether the original process $M$ is a POMDP or an MDP does not influence in any way the optimal solution of conformant planning. Note that the observation set $\Omega$ and the observation function $O$ are not used anywhere in the equations above.

Practical implementations of VI exploit the fact that the value function is always a piecewise linear convex function of the belief $x$. The functions $V_t^0(\cdot)$ and $Q_t^0(\cdot, a)$ are represented as finite sets of $\alpha$-*vectors*, each of them corresponding to a linear function of $x$. $V_t^0$ and $Q_t^0$ are then defined as the supremum (max) of the set of linear functions that represent them. All operations in equations (2) and (3) reduce to manipulation and production of $\alpha$-vectors. The sets of $\alpha$-vectors are regularly purged of vectors representing linear functions that are optimal nowhere in the belief space. Many algorithms differ only in the way they purge sets of $\alpha$-vectors. Although the belief space is continuous, all the computation is finite [10, 6].

The value function constructed when solving $M^0$ up to the planning horizon $H$ contains the expected reward of the best conformant plan in each possible initial belief state, and for each planning horizon less than or equal to $H$. To get the optimal plan for a particular starting belief $x_0$ (for instance, the certainty of being in a given state) and horizon $H$, we must simulate a trajectory by always executing the optimal action for the current belief state, which requires monitoring the belief state along the trajectory using equation (4). Since there is only one possible observation at each step, there is always only one possible belief at the next step. So, the trajectory can never branch.[1] We could as easily extract the optimal conformant plan for another starting belief and/or another planning horizon $h < H$. All the information that is important and hard to calculate is in the value function, which is computed only once. In OKP, we do not need to extract any plan before having reached the level $k$ where we decide to stop.

## 2.2　1-CONTINGENCY PLANNING

Similarly, the optimal 1-contingency plan is the optimal solution of a POMDP $M^1 = (S^1, A^1, \Omega^1, T^1, R^1, O^1)$. $M^1$ is constructed by duplicating $M^0$ and adding an *observe-and-branch* action between the two copies of $M^0$. Thus, each state $s \in S$ of the original POMDP $M$ is represented twice in $M^1$. One copy represents being in $s$ before the plan has branched, and the other represents being in $s$ after the plan has branched. The observe-and-branch action induces an irreversible transition from states of the first type to states of the second type. As for $k = 0$, the problem is completely non-observable, except that the observe-and-branch action allows making an ordinary observation as specified in the original POMDP $M$, and conditioning the next actions on this observation. If $M$ is an MDP, then the observe-and-branch action sees the current state exactly. Formally:

**States:** $S^1 = S \times \{0, 1\}$. The pair $(s, k)$, $s \in S$ and $k \in \{0, 1\}$, represents being in $s$ and having possibility of using the observe-and-branch action $k$ times in the future. Each $(s, 0)$ may be seen as an element of $S^0$, the state space of the conformant planning NOMDP $M^0$.

**Belief states:** The number of branch points that are still available for the future, $k$, is always known with certainty. All the uncertainty on the state $(s, k)$ of $M^1$ comes from the uncertainty on $s$. Therefore, a belief state for $M^1$ is a pair $(x, k)$ where $x$ is a probability distribution over $S$ and $k \in \{0, 1\}$.

**Actions:** $A^1 = A \cup \{a^{ob}\}$, where $a^{ob}$ is the observe-and-branch action. $a^{ob}$ is executable only in states $(s, 1)$, $s \in S$. $a^{ob}$ is a special *instantaneous* action: executing it does not increment time. As shown below, it can be used only once in each trajectory. The other actions $a \in A$ are called *ordinary* actions.

**Observations:** Formally, $\Omega^1 = \Omega$. However, useful observations can be made only through the observe-and-branch action $a^{ob}$. All other actions provide a non informative observation. To model this, we select arbitrarily one observation of the original process, we rename it $o^0$, and we use it to represent the non-informative observation produced by all actions different from $a^{ob}$. Observed after an ordinary action $a \in A$, $o^0$ means "I can't see anything interesting", and when it is observed after $a^{ob}$, it has the same semantics as in the original process $M$.

**Effects of ordinary actions:** The states $(s, 0)$, $s \in S$, represent an absorbing subset, that is, we cannot get out of this subset once we enter it (remember that only ordinary actions are possible in such states). All the transition probabilities, rewards and observation probabilities involving only such states are defined as in $M^0$. The only way to get out from states of type $(s, 1)$, $s \in S$, is through the observe-and-branch action. The transition probabilities, reward and observations involving only states of the type $(s, 1)$, $s \in S$, *and not the observe-and-branch action* $a^{ob}$, are also defined exactly as the transitions, rewards, and observations in $M^0$. That is: $T^1((s, k), a, (s', k)) = T(s, a, s')$, $R^1((s, k), a, (s', k)) = R(s, a, s')$, and $O^1(a, (s', k), o^0) = 1$, for all $(s, k, a, s') \in S \times \{0; 1\} \times A \times S$.

**Effect of the observe-and-branch action:** executing action $a^{ob}$ in state $(s, 1)$ leads with certainty to state $(s, 0)$, with the same number of time-steps to go. This action provides no reward and produces an observation following the observation probability of the original POMDP. Formally: $T^1((s, 1), a^{ob}, (s, 0)) = 1$, $R^1((s, 1), a^{ob}, (s, 0)) = 0$, and $O^1(a^{ob}, (s, 0), o) = O(s, o)$, for all $(s, o) \in S \times \Omega$.

The fact that the observe-and-branch action is instantaneous might make the solution of $M^1$ with VI look a little bit complicated *a priori*. However, it turns out that optimization over a finite horizon is straightforward. First, for all $x$ and all $t \leq H$, the value of belief state $(x, 0)$ at time $t$ in $M^1$ is equal to $V_t^0(x)$ in $M^0$. In other words, the result of the computation at level 0 (equations (1) through (3)) can be reused as is, it gives the value of each belief state $(x, 0)$ of $M^1$ at all $t \in \{0, 1, \ldots H\}$. Then, if we denote by $V_t^1(x)$ the value at time $t$ of belief $(x, 1)$ in $M^1$, then VI is summarized by the following equations:

$$V_H^1(x) = 0 \ , \tag{5}$$

and, for all $t \in \{0, 1, \ldots H - 1\}$:

$$V_t^1(x) = \max \left\{ Q_t^1(x, a^{ob}), \max_{a \in A} \left[ Q_t^1(x, a) \right] \right\} \ , \tag{6}$$

---

[1] It is also possible to simulate trajectories by following pointers from $\alpha$-vectors at time $t$ to $\alpha$-vectors at time $t + 1$ established when solving $M^0$, instead of maintaining the current belief. However, this technique appeared to be much slower in the context of OKP with $k > 0$, because it does not allow *not* building a branch for observations that are impossible *given the current belief* during plan extraction.

with

$$Q_t^1(x, a) = \left(\sum_{s \in S} x(s) R(s, a)\right) + \gamma V_{t+1}^1(\mathcal{B}_{o^0}^a(x)) \quad (7)$$

for all $a \in A$ (using equation (4) to calculate $\mathcal{B}_{o^0}^a(x)$), and

$$Q_t^1(x, a^{ob}) = \sum_{o \in \Omega} Q_t^1(x, a^{ob}, o) , \quad (8)$$

$$Q_t^1(x, a^{ob}, o) = \sum_{s \in S} x(s) O(s, o) V_t^0(\mathcal{B}_o^{a^{ob}}(x)) , \quad (9)$$

where $\mathcal{B}_o^{a^{ob}}(x)$ is the posterior belief after observing $o$, given by Bayes' rule:

$$\mathcal{B}_o^{a^{ob}}(x)(s') = \frac{x(s') O(s', o)}{Z} . \quad (10)$$

Note that if the original problem is an MDP, then equations (8) through (9) simplify as:

$$Q_t^1(x, a^{ob}) = \sum_{s \in S} x(s) V_t^0(x_s) , \quad (11)$$

where belief $x_s$ gives state $s$ with probability 1.

So, a practical solution of $M^1$ requires (i) having solved $M^0$ in advance; and (ii) one (backward) pass of VI through states $(s, 1), s \in S$, following equations (5) to (11). During the calculation of $V^1$, we read $\alpha$-vectors in the solution of $M^0$ to evaluate the observe-and-branch actions. Once the value function $V^1$ is calculated, we can extract the optimal 1-contingency plan for a given initial belief $x_0$ by simulating a trajectory in $M^1$. As long as the observe-and-branch action is not used, the trajectory may never branch. If at some point the $Q$-values $Q_t^1$ indicate that $a^{ob}$ is the optimal action for the current belief state, then a branch point is added to the plan. We must then calculate the posterior belief for each observation $o \in \Omega$ using equation (10) (that is, for each state $s \in S$ if $M$ is an MDP). Finally, the optimal branch for each $o$ is constructed by simulating a (non-branching) trajectory in $M^0$. Because $a^{ob}$ is not present in $M^0$, no more branch points can be added. Note that it may happen that the observe-and-branch action is never used during the travel through $M^1$. It shows that there exists a conformant plan that is at least as good as the best 1-contingency plan, so there is no need to use an observe-and branch action. Note also that, one more time, the optimal solution of $M^1$ contains the value of the best $k$-contingency plan for all $k \in \{0, 1\}$, all possible initial beliefs $x_0$, and all planning horizons less than or equal to $H$.

## 2.3 BALANCED $k$-CONTINGENCY PLANNING

In general, the $k$-contingency planning problem ($k \geq 2$) may be modelled as a POMDP $M^k$ built on $M^{k-1}$ by adding a copy of $S^0$ connected to the $(k - 1)^{\text{th}}$ level of $M^{k-1}$ by the observe-and-branch action. All the equations of the previous section can be re-used by replacing superscript 1 by $k$ and superscript 0 by $k - 1$. That is:

$$V_H^k(x) = 0 , \quad (12)$$

$$V_t^k(x) = \max \left\{ Q_t^k(x, a^{ob}), \max_{a \in A} \left[Q_t^k(x, a)\right] \right\} , \quad (13)$$

$$Q_t^k(x, a) = \left(\sum_{s \in S} x(s) R(s, a)\right) + \gamma V_{t+1}^k(\mathcal{B}_{o^0}^a(x)) , \quad (14)$$

$$Q_t^k(x, a^{ob}) = \sum_{o \in \Omega} Q_t^k(x, a^{ob}, o) , \quad (15)$$

$$Q_t^k(x, a^{ob}, o) = \sum_{s \in S} x(s) O(s, o) V_t^{k-1}(\mathcal{B}_o^{a^{ob}}(x)) . \quad (16)$$

If the solution of $M^{k-1}$ is known, then the solution of $M^k$ requires only one pass of VI through states at level $k$ (that is, states $(s, k), s \in S$), reading $\alpha$-vectors in $V_t^{k-1}$ to evaluate the observe-and-branch action. Once the value functions $V_t^k$ are determined, we can easily extract the best (balanced) $k$-contingency plan for a given initial belief by simulating a trajectory in $M^k$. When the observe-and-branch action is used, the trajectory branches and one branch for each possible observation $o \in \Omega$ must be built by simulating a trajectory in $M^{k-1}$. This is why the algorithm produces *balanced* contingency plans: at each branch point at level $l \leq k$, *each exiting branch* (which is in fact a tree) may contain up to $l - 1$ branch points (equation (16)). Therefore, each trajectory through the plan tree may traverse up to $k$ branch points. As previously, the algorithm does not have to use all the branch points allowed if there is no utility to be gained by doing so. Therefore, the version of OKP presented in this section produces an optimal plan with *at most* $k$ branch points in each trajectory.[2]

## 3 EXTENSIONS

OKP may easily be adapted to other variants of the limited contingency planning problem.

### 3.1 TYPES OF PLANS

First, the algorithm can search for other types of plans. For instance, we can search for the optimal linear $k$-contingency plan as defined in Section 1.1, that is, the best plan with (at most) $k$ branch points, all of them on one trajectory through the plan. In this case, each level $l \in \{1, 2, \ldots k\}$ of $M^k$ contains $|\Omega|$ observe-and-branch actions, $\{a_o^{ob}, o \in \Omega\}$. The semantics of $a_o^{ob}$ is "observe, branch, and use the $l - 1$ remaining branch points in the branch associated with observation $o$". Equation (13) becomes

$$V_t^k(x) = \max\left\{\max_{o \in \Omega}\left[Q_t^k(x, a_o^{ob})\right], \max_{a \in A}\left[Q_t^k(x, a)\right]\right\} ,$$

where

$$Q_t^k(x, a_o^{ob}) = Q_t^{k-1}(x, a_o^{ob}, o) + \sum_{o' \in \Omega \setminus \{o\}} Q_t^0(x, a_o^{ob}, o') .$$

---

[2] Note that the plan extraction phase of this version of OKP is exponential in $k$. This is an artifact due to the particular variant of the problem addressed. What we call a "balanced $k$-contingency" plan contains in fact a number of branch points exponential in $k$. Therefore, extracting such a plan from the solution of the POMDP is exponential in $k$. This is not the case for the other variants of the algorithm presented in Section 3.1.

Similarly, we can tackle the general $k$-contingency planning problem (at most $k$ branches over the whole plan without any other constraint), by adding multiple observe-and-branch actions at each level of $M^k$. Here we must model one observe-and-branch action for each possible way to distribute the $k - 1$ remaining branch points in the $|\Omega|$ exiting branches. Therefore, the number of different observe-and-branch actions required at level $k$ is

$$\frac{(|\Omega| + k - 2)!}{(|\Omega| - 1)!(k - 1)!} .$$

So this variant of OKP is somewhat impractical for large $k$. As shown below, a way to limit the complexity of the algorithm is to change the branch conditions.

### 3.2 BRANCH CONDITIONS

The algorithm of Section 2 creates one particular branch for each observation $o \in \Omega$ that can possibly be made after the observe-and-branch action (although it considers only the observations that are possible given the current belief during plan extraction). In other words, there may be up to $|\Omega|$ branches stemming from each branch point of the plan. In some variants of the limited contingency planning problem, we may want to limit the number of branches exiting from each branch point by grouping several observations together.

OKP can be adapted to any kind of branch condition. For instance, if we want the plan to use binary branch points, then we must create one observe-and-branch action $a_{\Omega'}^{ob}$ for each possible way to partition the observation set $\Omega$ into two non-empty subsets $\Omega'$ and $\Omega \setminus \Omega'$. Equation (13) becomes

$$V_t^k(x) = \max\left\{\max_{\Omega'}\left[Q_t^k(x, a_{\Omega'}^{ob})\right], \max_{a \in A}\left[Q_t^k(x, a)\right]\right\} ,$$

$$Q_t^k(x, a_{\Omega'}^{ob}) = Q_t^k(x, a_{\Omega'}^{ob}, \Omega') + Q_t^k(x, a_{\Omega'}^{ob}, \Omega \setminus \Omega') ,$$

where

$$Q_t^k(x, a_{\Omega'}^{ob}, \Omega') = \Pr(\Omega' \mid x) V_t^{k-1}(\mathcal{B}_{\Omega'}^{a_{\Omega'}^{ob}}(x)) ,$$

$$\Pr(\Omega' \mid x) = \sum_{s \in S} x(s) \sum_{o \in \Omega'} O(s, o) ,$$

$$\mathcal{B}_{\Omega'}^{a_{\Omega'}^{ob}}(x)(s') = \frac{x(s') \sum_{o \in \Omega'} O(s', o)}{Z} ,$$

and similarly for $Q_t^k(x, a_{\Omega'}^{ob}, \Omega \setminus \Omega')$. Note that there are $2^{|\Omega|} - 2$ such actions (subsets $\Omega'$), which is a considerable number in most cases.

The equations above correspond to balanced $k$-contingency planning. If we are looking for other types of plans, then we must create a different observe-and-branch action for each possible branch condition *and* each possible way of distributing the remaining branch points in the stemming branches. However, the number of ways of distributing branch points is greatly reduced (compared to the formulas of Section 3.1) when we use compact branch conditions. For instance, if we look for the optimal plan with at most $k$ binary branch points overall, then there are $2^{|\Omega|} - 2$ different branch conditions, but only $k$ ways to distribute the $k - 1$ remaining branch points in the two exiting branches. Therefore, the total number of observe-and-branch actions at level $k$ is $(2^{|\Omega|} - 2)k$.

The computational price of compact branch conditions can be greatly reduced in the particular case where the observation $o$ represents a numerical value.[3] In this case, we can focus the search on a particular kind of branch condition based on threshold. Each branch point is defined by a threshold $o^T \in O$. There are two exiting branches: one corresponds to observing a value $o \in O$ less than or equal to $o^T$, and the other corresponds to values greater than $o^T$. Thus, the total number of different branch conditions is $|\Omega| - 1$. As there are only two exiting branches, there are only $k$ ways to distribute the remaining branch points. Therefore, the total number of observe-and-branch actions at level $k$ of the strict $k$-contingency planning POMDP is only $(|\Omega| - 1)k$.

### 3.3 GENERAL POMDPS

Finally we can relax the hypothesis on the observation probabilities of the original POMDP $M$. In Section 2, we assumed that the observation probabilities depend only on the arrival state $s'$ (that is, $O(s', o)$), while the general formalism of POMDPs assumes that they also depend on the last action $(O(a, s', o))$, which allows a richer model of sensory actions. The problem is that, when we move to this more general framework, the observation probabilities of $a^{ob}$ in $M^k$, previously defined as $O^k(a^{ob}, (s, k-1), o) = O(s, o)$, are not well defined anymore. The observation following the use of the observe-and-branch action depends on the action performed at the previous time step, which violates the (first order) Markov property.

---
[3]Actually, it is not necessary that the observation is a numerical variable. It is sufficient that there be a complete order defined over it.

One way to deal with this situation is to introduce the last action executed into the Markov state of $M^k$. Another, equivalent, way to model this is as follows: instead of adding $N_k$ observe-and-branch actions to the preexisting $|A|$ actions at each level $k$ (where $N_k$ is the total number of branch conditions and ways of distributing $k - 1$ remaining branch points in the exiting branches), we create $N_k$ (new) copies of each action $a \in A$. Each copy corresponds to executing $a$, and then branching the plan following the protocol of a particular observe-and-branch action. For instance, in the case of balanced $k$-contingency planning with $|\Omega|$-ary branch points (as in Section 2), we duplicate each action $a \in A$ and call $\bar{a}$ its copy ($\bar{A}$ is the set of all copies). $\bar{a}$ represents executing $a$, not discarding the resulting observation, and branching the plan based on this observation following the protocol of action $a^{ob}$ of Section 2. The equations of VI become:

$$V_t^k(x) = \max \left\{ \max_{a \in A} \left[Q_t^k(x,a)\right], \max_{\bar{a} \in \bar{A}} \left[Q_t^k(x,\bar{a})\right] \right\} ,$$

$$Q_t^k(x, \bar{a}) = \sum_{o \in \Omega} Q_t^k(x, \bar{a}, o) ,$$

$$Q_t^k(x, \bar{a}, o) = \sum_{s \in S} x(s)O(s,o)\left(R(s,a) + \gamma V_{t+1}^{k-1}(\mathcal{B}_o^{\bar{a}}(x))\right) ,$$

$$\mathcal{B}_o^{\bar{a}}(x)(s') = \frac{x(s')O(a,s',o)}{Z} .$$

Note that we are not concerned with this issue if the original process $M$ is a fully observable MDP.

## 4 EXPERIMENTS

We implemented OKP using Cassandra's POMDP solver available on the Internet.[4] We used the witness algorithm [10] to solve OKP's multiple level POMDP. The results presented in this paper concern the variant of OKP that searches for balanced contingent plans (Section 3.1), building a branch for each possible observation (Section 3.2). We focus on two simple test bed problems.

As Hyafil and Bacchus stressed for the particular case $k = 0$, OKP for general $k$ is able to prune the plan space (using Bellman's optimality principle), but it computes (the value of) the optimal plan in every belief state at every horizon, while we may be interested only in a single initial belief and the belief states reachable from it. To measure the value of this trade-off, we implemented in the same environment as OKP, an algorithm that systematically searches and evaluates all possible contingent plans for a given $k$, horizon, and initial belief, taking into account only reachable belief states. Its performance gives an indication of the size of the search space, and how OKP is able to prune the search using Bellman's optimality principle.

The first problem we used is a variant of the tiger problem [10]. In this problem, the agent is standing in front of two doors (*left* and *right*). Behind one door lies a dangerous tiger, and there is a reward behind the other door. Therefore, there are two different world states: *tiger–left* and *tiger–right*. The initial position of the tiger is unknown, and the initial probability on the tiger position is uniform over the two doors. The agent has three possible actions: opening one of the doors (*open–left* and *open–right*), or listening to try to guess where the tiger is (*listen*). The *listen* action does not change the state of the world, it costs 1 unit of utility, and provides a noisy observation that can take two possible values: *hear–tiger–left* and *hear–tiger–right*. If the state of the world is *tiger–left*, then the probability of observing *hear–tiger–left* is 0.85 and the probability of observing *hear–tiger–right* is 0.15. Similarly, the probability of hearing the tiger to the right when the tiger is actually to the right is 0.85. Opening the door behind which the tiger lies provides a "reward" of -10. Opening the other door brings a reward of +6. After opening a door, the problem is reset in its original state (that is, the agent is brought back in front of the doors and the new position of the tiger is drawn at random uniformly). Given these parameters, the optimal conformant plan over a horizon of $H$ time-steps is to *listen* $H$ times. At each step, it provides the reward −1 with certainty, while opening an arbitrary door (we are not allowed to condition the choice of the door on the result of previous *listen* actions) brings the expected reward: 0.5 (-10) + 0.5 (6) = -2. The discount factor is set to 1 (no discount).

We ran OKP and plan enumeration on the tiger problem for different planning horizons $H$ and levels $k$. Fig. 1 shows the optimal contingent plans obtained with a sample of small values for $H$ and $k$. Fig. 2 shows the evolution of the value of the optimal contingent plan as a function of $k$ and $H$. Finally, Fig. 3 shows the evolution of the total time taken by the algorithm as a function of $k$ and $H$. These results clearly show the exponential blow-up of the search space and how OKP is able to resist it by efficiently pruning the search. In this example, Bellman's optimality principle allows a drastic reduction in the complexity of the search that largely compensates for the fact that we have to deal with (belief) states that are unreachable.

The second problem is a small maze world due to Horstmann and Zilberstein [8] and represented in Fig. 4. In this problem, the agent starts from the location marked with an S and must end-up in the goal location G. The agent can use 4 actions, N, S, E and W, that allow it to move 1 or 2 positions in the desired direction with equal probability (unless a wall blocks the way). The goal state is absorbing. The observation available (when we decide to branch) is the presence or absence of a wall on each side of the square that defines the agent's location. Thus, there are 8 different possible observations (and 11 states). The agent gets a zero reward at every step except when it enters the goal state. Therefore, there is no time pressure on the agent: it does not get a bigger reward for getting to the goal earlier, and it must simply maximize its probability of reaching the goal inside of the planning horizon. Fig. 4 contains an example of an optimal contingent plan for this problem. Fig. 5 and 6 show the evolution of the value of the optimal plan and of the execution time of the two algorithms on this problem. As for the previous example, the trade-off adopted by OKP is highly valuable.

---

[4] http://www.cs.brown.edu/research/ai/pomdp/

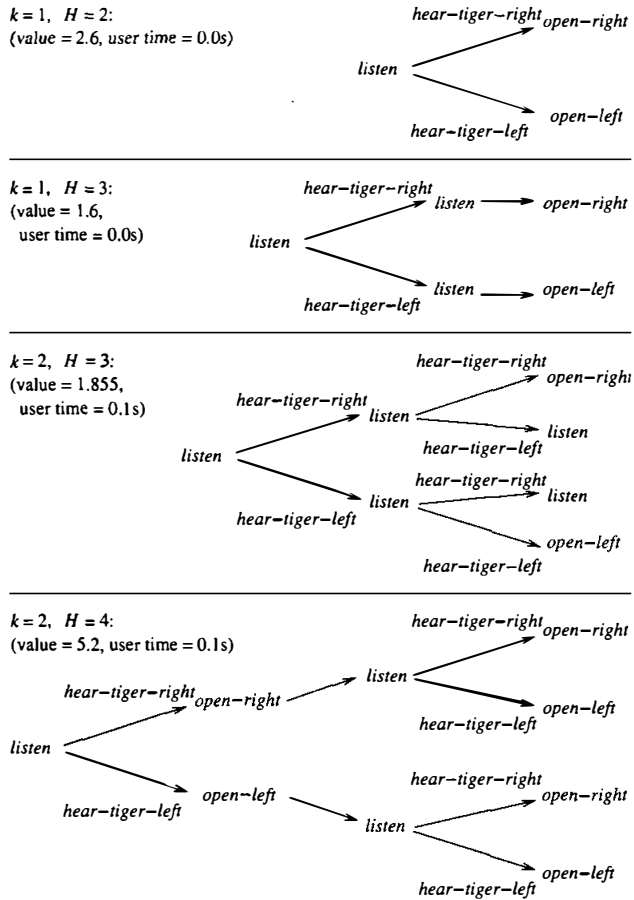

Figure 1: Optimal contingent plans for the tiger problem.

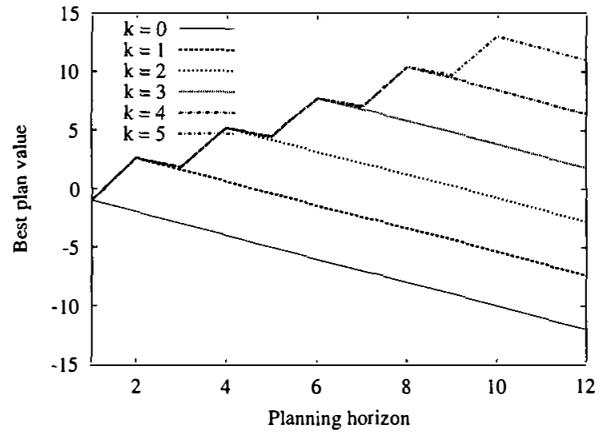

Figure 2: Value of the optimal contingent plans of the tiger problem.

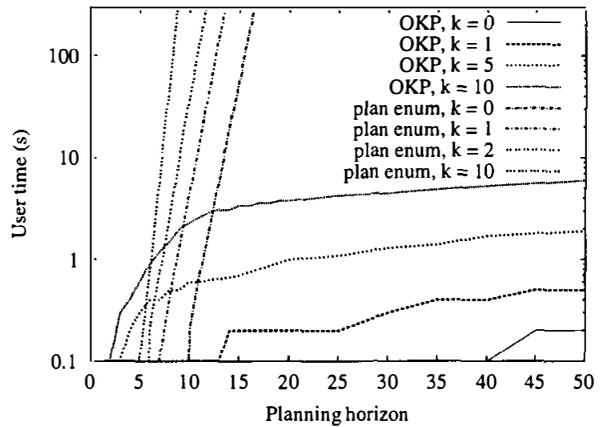

Figure 3: Execution time of OKP and plan enumeration in the tiger problem.

Finally, we experimented on the GRID-10×10 problem designed by Hayfil and Bacchus [9] to show the limits of the POMDP approach to conformant planning. This problem is constituted of an empty 10×10 square room. The goal state is a corner of the room and the start state state is a fixed location in the middle of the room. The four actions available, N, S, E, and W, allow the agent to move of one position in the grid, but there is noise in the direction of this move. The actions N and S move the agent in the designated direction with probability 0.9, and to the West and East directions with probability 0.05 each. Similarly, the E and W action succeed with probability 0.8 and move the agent to the North and South with probability 0.1. As in Horstmann and Zilberstein's maze, the agent can perceive only nearby walls. The algorithms execution time for this problem is presented in Figure 7. These results are consistent with Hyafil and Bacchus's. They show that the plan enumeration technique is faster than OKP in this particular problem. This may be explained by observing that, for small values of the planning horizon, there are much less reachable states than the total number of states. Therefore, the reachability analysis of the plan enumeration algorithm allows saving more time than Bellman's optimality principle buys us in OKP. It suggest that the best algorithms will be obtained by combining reachability analysis and Bellman's optimality principle.

## 5 RELATED WORK

A number of probabilistic contingency planning systems have been developed that can deal with partial observabil-ity, including C-Buridan [7], DTPOP [14], Mahinur [13], P-Graphplan [3], C-MAXPLAN [12], ZANDER [12], and heuristic search through the belief space [4, 2]. Since the limited contingency planning problem may be modelled as a POMDP, all of them can potentially be applied to this problem. In a sense, the contribution of this paper is to show how to cast the limited contingency planning problem as a problem of planning with partial observability. Not all of these systems attempt to maximize the expected reward. For instance, the objective for many of them is to find a plan with a success probability exceeding a given threshold. They can potentially be used to find a limited contingency plan that succeeds with a minimum probability. Also, by raising the probability threshold, one could in theory force any of these systems to continue searching for an optimal plan or policy. We believe that it should be relatively easy to do this for the partial-order planners C-Buridan [7], DTPOP [14], and Mahinur [13]. For these systems, all that would be required is to incorporate a counter into the planning algorithm so that no more than $k$ branches could be added to the plan. For C-MAXPLAN [12] and ZANDER [12] one could write exclusion axioms that prohibit more than $k$ observation axioms from appearing in the plan. However, if there are $n$ possible observations, $\binom{n}{k+1}$ exclusion axioms would be required. Finally, heuristic search through the belief space [4, 2] can be applied directly to the POMDP $M^k$ of $k$-contingency planning. It amounts to in-troducing the number of branch points remaining as a fully observable component of the state.

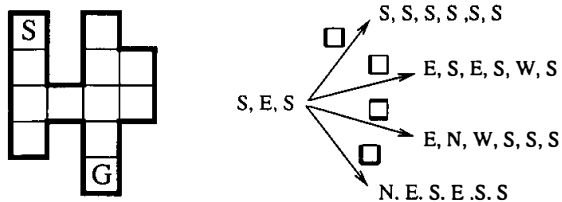

Figure 4: Horstmann and Zilberstein's maze problem and the optimal contingent plan for $k = 1$ and $H = 9$.

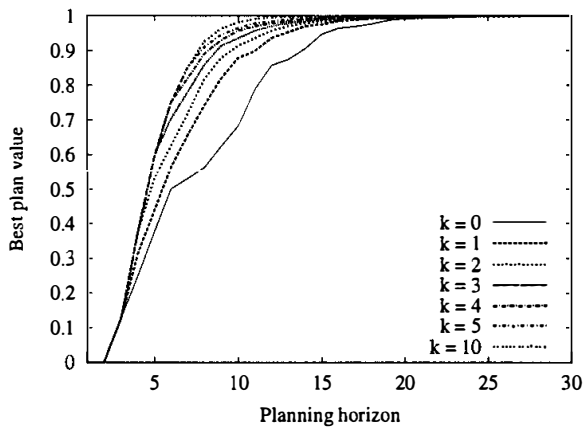

Figure 5: Value of the optimal contingent plans in Horstmann and Zilberstein's maze.

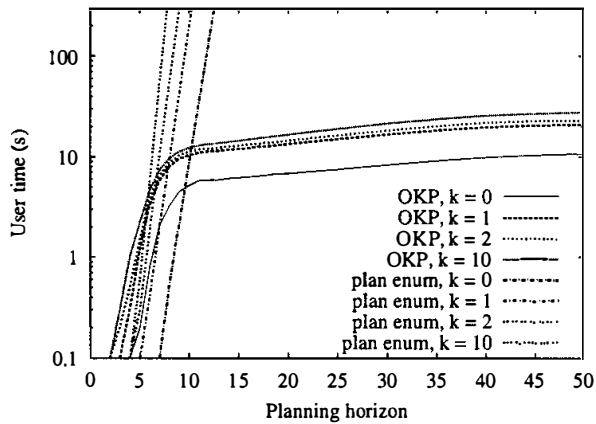

Figure 6: Execution time of OKP and plan enumeration in Horstmann and Zilberstein's maze.

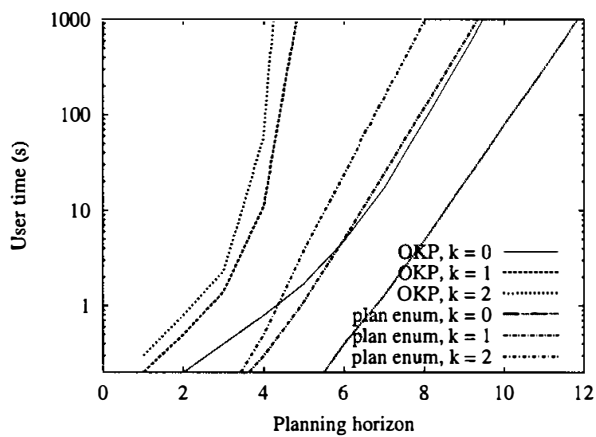

Figure 7: Execution time of OKP and plan enumeration in the GRID-10x10 problem.

## 6 CONCLUSIONS

We presented OKP, a new algorithm that is able to find optimal solutions to a variety of $k$-contingency planning problems. The basic principle of OKP is to recognize that the belief state borrowed from POMDPs contains all the information necessary to allow a DP solution to limited contingency planning. We have shown experimentally that the time gained by pruning the plan space using Bellman's optimality principle may largely compensates for the fact that we have to deal with (belief) states that are unreachable, but that this trade-off is not be beneficial in all cases. This work, as well as some recent work on conformant planning, shows that Bellman's optimality principle is a powerful tool for many *optimal* planning problems (where we have to find the best plan over a set plans), not just search-ing for the optimal policy. By showing how to cast the limited contingency planning problem as a problem of planning with partial observability, this work allows the application of many previous algorithms to limited contingency planning.

**Acknowledgments**

We thank Richard Dearden and Sailesh Ramakrishnan for comments on the material, and Rich Washington for helpful feedback on the paper. This work was supported by the NASA Intelligent Systems Program.

# References

[1] P. Bertoli, A. Cimatti, and M. Roveri. Heuristic search + symbolic model checking = efficient conformant planning. In *Proceedings of the Seventeenth International Joint Conference on Artificial Intelligence*, 2001.

[2] P. Bertoli, A. Cimatti, M. Roveri, and P. Traverso. Planning in mondeterministic domains under partial observability via symbolic model checking. In *Proceedings of the Seventeenth International Joint Conference on Artificial Intelligence*, 2001.

[3] A. Blum and J. Langford. Probabilistic planning in the Graphplan framework. In *Proceedings of the Fifth European Conference on Planning*, 1999.

[4] B. Bonet and H. Geffner. Planning with incomplete information as heuristic search in belief space. In *Proceedings of the Fifth International Conference on Artificial Intelligence Planning and Scheduling*, pages 52–61, 2000.

[5] J. Bresina, R. Dearden, N. Meuleau, S. Ramakrishnan, D. Smith, and R. Washington. Planning under continuous time and resource uncertainty: A challenge for AI. In *Proceedings of the Eighteenth Conference on Uncertainty in Artificial Intelligence*, 2002.

[6] A.R. Cassandra, M.L. Littman, and N.L. Zhang. Incremental Pruning: A simple, fast, exact method for partially observable Markov decision processes. In *Proceedings of the Thirteenth Conference on Uncertainty in Artificial Intelligence*, pages 54–61, San Francisco, CA, 1997. Morgan Kaufmann.

[7] D. Draper, S. Hanks, and D. Weld. Probabilistic planning with information gathering and contingent execution. In *Proceedings of the Second International Conference on Artificial Intelligence Planning and Scheduling*, pages 31–36, 1994.

[8] M. Horstmann and S. Zilberstein. Automated generation of understandable contingency plans. In *ICAPS-03: Proceedings of the Workshop on Planning under Uncertainty and Incomplete Information*, 2003.

[9] N. Hyafil and F. Bacchus. Conformant probabilistic planning via CSPs. In *Proceedings of the Thirteenth International Conference on Automated Planning and Scheduling*, 2003. To appear.

[10] L.P. Kaelbling, M.L. Littman, and A.R. Cassandra. Planning and acting in partially observable stochastic domains. *Artificial Intelligence*, 101:99–134, 1998.

[11] M. Littman, J. Goldsmith, and M. Mundhenk. The computational complexity of probabilistic planning. *Journal of AI Research*, 9:1–36, 1998.

[12] S. Majercik and M. Littman. Contingent planning under uncertainty via stochastic satisfiability. In *Proceedings of the Sixteenth National Conference on Artificial Intelligence*, 1999.

[13] N. Onder and M. Pollack. Conditional, probabilistic planning: A unifying algorithm and effective search control mechanisms. In *Proceedings of the Sixteenth National Conference on Artificial Intelligence*, pages 577–584, 1999.

[14] M. Peot. *Decision-Theoretic Planning*. PhD thesis, Dept. of Engineering-Economic Systems, Stanford University, 1998.

[15] M.L. Puterman. *Markov Decision Processes: Discrete Stochastic Dynamic Programming*. Wiley, New York, NY, 1994.